\documentclass[10pt,journal,compsoc]{IEEEtran}

\ifCLASSOPTIONcompsoc
  \usepackage[nocompress]{cite}
\else
  \usepackage{cite}
\fi

\ifCLASSINFOpdf
\usepackage{graphicx}
\usepackage{cite}
\usepackage{tikz}
\usepackage{comment}
\usepackage{amsmath,amssymb} 
\usepackage{color}
\usepackage{caption}
\usepackage[accsupp]{axessibility}  %

\usepackage{hyperref}
\usepackage{makecell}
\usepackage{amssymb}
\usepackage{amssymb}
\usepackage{pifont}
\usepackage{mathtools}
\usepackage{lipsum}

\newcommand{\prob}{multi-object tracking}

\newcommand{\cmark}{\ding{51}}%
\newcommand{\xmark}{\ding{55}}%
\DeclareMathOperator*{\argmax}{argmax} 
\graphicspath{ {Images/} }

\begin{document}
\raggedbottom

\title{
	UnsMOT: Unified Framework for Unsupervised Multi-Object Tracking with Geometric Topology Guidance
}

\author {
		Son Tran, Cong Tran, Anh Tran, Cuong Pham,~\IEEEmembership{Member,~IEEE} \\%
\IEEEcompsocitemizethanks{
\IEEEcompsocthanksitem S. Tran is with the Post of Computer Science, Posts and Telecommunications Institute of Technology, Hanoi 100000, Vietnam.\protect\\
E-mail: tthanhson1199@gmail.com.
\IEEEcompsocthanksitem C. Tran is with the Department of Information Technology, Posts and Telecommunications Institute of Technology, Hanoi 100000, Vietnam.\protect\\
E-mail: congtt@ptit.edu.vn.

\IEEEcompsocthanksitem A. Tran is with the Computer Vision department, VinAI Research, Hanoi 100000, Vietnam. 
\protect\\
E-mail: v.anhtt152@vinai.io.

\IEEEcompsocthanksitem C. Pham is with the Department of Information Technology, Posts and Telecommunications Institute of Technology and is also with the Computer Vision department, VinAI Research, Hanoi 100000, Vietnam. 
\protect\\
E-mail: v.anhtt152@vinai.io.

(Corresponding author: Cuong Pham.)}}%

\markboth{Submitted to IEEE Transactions on  Pattern Analysis and Machine Intelligence}{}

\IEEEtitleabstractindextext{
\begin{abstract}
Object detection has long been a topic of high interest in computer vision literature. Motivated by the fact that annotating data for the multi-object tracking (MOT) problem is immensely expensive, recent studies have turned their attention to the {\em unsupervised} learning setting. In this paper, we push forward the state-of-the-art performance of unsupervised MOT methods by proposing UnsMOT, a novel framework that explicitly combines the appearance and motion features of objects with geometric information to provide more accurate tracking. Specifically, we first extract the appearance and motion features using CNN and RNN models, respectively. Then, we construct a graph of objects based on their relative distances in a frame, which is fed into a GNN model together with CNN features to output geometric embedding of objects optimized using an unsupervised loss function. Finally, associations between objects are found by matching not only similar extracted features but also geometric embedding of detections and tracklets. Experimental results show remarkable performance in terms of HOTA, IDF1,  and MOTA metrics in comparison with state-of-the-art methods.

\end{abstract}

\begin{IEEEkeywords}
Graph Neural Network, Multi-Object Tracking,  Unsupervised Learning
\end{IEEEkeywords}}
\maketitle

\IEEEdisplaynotcompsoctitleabstractindextext

%
\IEEEpeerreviewmaketitle

\begin{IEEEkeywords}
Graph Neural Network, Multi-Object Tracking,  Unsupervised Learning
\end{IEEEkeywords}
\IEEEpeerreviewmaketitle


\section{Introduction}
The multi-object tracking (MOT) problem aims at locating all occurrences of multiple objects in a video and following each object across the next frames as long as the objects are still observable in the current frame. In this problem, it is well-known that annotating data for training is extremely time-consuming and expensive. For example, creating training labels for a six-minute video in the MOT15 Challenge takes approximately 22 hours of human effort \cite{pathtrack}. Annotating 26 hours of the VIRAT dataset \cite{oh2011large} costs thousands of dollars using state-of-the-art procedures \cite{oh2011large,vondrick2011video}. Recently, a topic rising in popularity is to unsupervisedly tackle the MOT problem, which results in several proposed methods, including UnsupTrack \cite{simpleuns} and UNS20regress \cite{uns20}. In these studies, the appearance and motion features of objects have been well-captured using delicately designed convolution neural network (CNN) and recurrent neural network (RNN) models.

A common approach is to  apply an object detector on each frame and then associate resulting detections to tracks  \cite{gsdt,gcnnmatch,gnnmot,leal2016learning}, the so-called tracking-by-detection paradigm. Recent methods usually utilize deep neural networks like CNN and RNN to extract features for solving the association problem, but they largely ignore the relationship between objects. Recent methods like   \cite{gcnnmatch,gnnmot,gsdt,gsmtrack} attempts to address this by modeling the interaction between objects using graph neural networks (GNN),  showing considerably substantial gains in terms of performance. However,  it is not straightforward to train such a GNN model without video-level labels. As such, there have not been many attempts to apply GNN to the unsupervised MOT setting.

Motivated by the intuition that the topological information between objects in the same frame (intra-frame topological information) can also be utilized to unsupervisedly aid in associating objects, we design a novel MOT framework that pushes forward the performance of unsupervised SOTA methods by combining appearance, motion, and intra-frame topological information of objects. To this end, we first utilize CNN and RNN models to capture the appearance and motion features of objects. Next, our important contribution to unsupervised \prob{} is combining the appearance features with the geometric information to form a graph of objects within the same frame. Then, the graph is fed into a GNN model to extract a latent representation of intra-frame topological information. Finally, we provide the association of objects between the current and previous frames using a weighted sum of similarity measures.  
We perform extensive experiments to validate the superiority of our proposed framework by making a performance comparison of UnsMOT with eight other benchmark methods, including both unsupervised and supervised ones, on three datasets from the MOT challenge, namely MOT16, MOT17, and MOT20. Experimental results show that our framework consistently outperforms all unsupervised benchmark approaches in terms of three well-applied performance metrics, namely HOTA, and IDF1, on both datasets. Interestingly, our unsupervised approach outperforms even recent supervised benchmark methods. Additionally, we provide an ablation study to comprehensively validate each component's effects in UnsMOT.
\section{Related Work}
\subsection{Supervised Multi-Object Tracking}
Supervised multi-object tracking methods rely on a large corpus of video-level bounding boxes and annotated datasets. The most dominant paradigm for object tracking is tracking-by-detection, where first an off-the-shelf object detector is used to detect objects in a frame, then these detections are associated with existing active tracklets up to the current frame \cite{yu2016poi,leal2016learning,sun2019deep,bae2017confidence}. Another line of work focuses on using Re-ID models \cite{ye2021deep} to match objects between frames or different cameras like \cite{yan2019learning,simpleuns,munjal2019query,han2019re}. \cite{yan2019learning} utilizes information present between objects to increase the identification confidence using a graph convolutional network. Given a target object in an image, they define context candidates as the remaining objects within the same image. Noisy pairs of context candidates from both the probe and gallery images are filtered using a relative attention model. To determine if a target pair of objects corresponds to the same identity, the authors proposed constructing a graph, with nodes consisting of the remaining unfiltered context pairs and the target pair and edges connecting a target pair and every context pair. Then, the model learns to output the similarity of the target pair.

Aside from the tracking-by-detection paradigm, several methods extended object detection networks to also regress each object's bounding box to the next frame or output an embedding to identify each object in the next frame  \cite{centertrack,jde,gsdt,tracktor,shuai2021siammot,pang2021quasi,wu2021track} by extending existing object detection networks like \cite{duan2019centernet,ren2015faster}, allowing the detection and association models to be optimized in an end-to-end manner, thus potentially boosting performance compared to tracking-by-detection methods in which each network is optimized independently. 
\subsection{Unsupervised Multi-Object Tracking}
Before the flourish of deep learning, unsupervised methods such as SORT \cite{sort} and V-IOU \cite{iou} associated detections by using handcrafted spatial and visual cues like a Kalman filter or optical flow. More recent unsupervised methods like \cite{uns20,simpleuns,vo2020self} adopted one or several deep learning models, leading to significantly better performances. A common approach for these unsupervised models is to enforce the model to learn useful features by solving a pretext task, then adapt these features to the target task with little human interaction, which is termed self-supervised learning. Specifically, SimpleUns \cite{simpleuns} automatically creates noisy labels for tracks in the video and then trains a ReID CNN model to predict this label given any detection from any tracks. UNS20regress \cite{uns20} instead optimizes the consistency of the score matrix outputted from a CNN and an RNN. The models optimized in this manner are forced to learn relevant visual and motion data respectively to maintain high consistency between their outputs.
\subsection{Multi-Object Tracking Based on Graph Neural Networks}
Recently, there have been multiple attempts to model MOT using graphs to boost the performance of supervised MOT methods. In GCNNMatch and GSDT\cite{gcnnmatch,gsdt}, the authors employed a graph convolution layer to model the interaction between objects and demonstrated a significant increase in the tracking performance, suggesting that relations between objects are beneficial for the task. In GCNNMatch\cite{gcnnmatch}, the authors proposed a method combining GNN model with Sinkhorn algorithm for end-to-end learning of an affinity matrix between detections and tracklets. Similar to GCNNMatch\cite{gcnnmatch}, GMTracker \cite{gmtracker} utilized a quadratic programming differentiable operation in combination with a cross graph GNN to acquire the affinity matrix, where a fully-connected graph was constructed for each frame independently, in which each node represented an object.  Different from previously mentioned methods, a novel graph construction procedure that treated every pixel of the image as a node to jointly optimize an object detection and an association model was proposed in GSDT \cite{gsdt}.  In parallel, \cite{braso2020learning} instead construct a graph connecting every object in one frame to another. The authors then proposed a novel message passing network to propagate information from nodes and edges for a fixed number of iterations. This information is used to predict active edges between nodes.
\subsection{Discussion}
While achieving highly accurate performance on many benchmark datasets, supervised methods require expensive and laborious labeling. Notably, recent unsupervised methods have outperformed several supervised methods \cite{uns20,simpleuns}.  However, the relationships between objects, which have been demonstrated to help improve performance \cite{gcnnmatch,gsdt}, have not been taken into account in recent studies under the unsupervised setting. 
\section{Methodology}
\subsection{Problem Definition}
Multi-object tracking is the problem of detecting and following multiple objects of interest across frames of a given video. In this study, we follow the tracking-by-detection paradigm as in recent works \cite{gcnnmatch,gsdt,uns20}. Specifically, at each frame, we are given a set of $n$ objects of interest, termed detections denoted as \(\mathcal{D}=\{d_0, d_1, \cdots, d_n\}\), and the set of $m$ active tracklets corresponding to historical detections, denoted as $\mathcal{T} = \{t_0, t_1, \cdots, t_m\}$. Each detection $d_i$ consists of two components, including: 1) an object bounding box $b_i^\mathcal{D}$ and 2) an image region cropped by $b_i^\mathcal{D}$, denoted as $img_i^\mathcal{D}$. Each tracklet $t_i$ represents the last observed instance of an object in previous frames, which also consists of a  bounding box $b_i^\mathcal{T}$ and a cropped image $img_i^\mathcal{T}$. \footnote{Similarly to many recent studies \cite{gcnnmatch,uns20,leal2016learning}, we assume that the detection problem is solved using a pre-trained object detector, thus there is no conflict with our claim on an unsupervised setting.}

Our study aims to {\em unsupervisedly} associate each $d_i \in \mathcal{D}$ with at most one $t_j \in \mathcal{T}$ such that $d_i$ and $t_j$ belong to the same ground truth object. To this end, we first calculate an affinity score $s_{i, j}$ between each pair of $d_i$ and $t_j$. We want to associate a detection $d_i$ with at most one $t_j$ such that their affinity score $s_{i,j}$ is the highest among all pairs $(i,j)$. Formally, let $S^A, S^M$, and $S^G$ denote the appearance, motion, and topological similarity matrices, respectively. The size of all the three matrices is $n \times m$, where each element at the $i$-th row and $j$-th column represents the respective similarity score of $d_i$ and $t_j$ according to each of the three aspects. The detailed implementation of the similarities will be explained in detail in Section \ref{sec:proposed_framework}. The final affinity matrix $ S = \{s_{i,j}\}$ is calculated as:
\begin{equation}
s_{i, j} = C(s^A_{i, j}, s_{i, j}^M, s_{i, j}^G)
\end{equation}
where $C(\cdot)$ is a combination function that combines $s^A_{i, j}, s_{i, j}^M$, and $s_{i, j}^G$ into a single similarity score $s_{i, j}$.
We denote  the binary association matrix $\mathcal{A}=\{a_{i,j}\}$,  with $a_{i,j} \in \{0,1\}$ and $a_{i,j} = 1$ indicating that a detection $d_i$ is associated with a tracklet $t_j$. We aim to find the optimal assignment matrix ${\mathcal{A}}^*$ by solving an optimization problem as follows:

\begin{align}
\label{eq:1}
\mathcal{A}^* = \argmax_{\mathcal{A}} \quad & \sum_{i=1}^n \sum_{j=1}^m s_{i,j} \times {a}_{i,j}, \\
\text{subject to}\quad 
& a_{i,j} \in \{0,1\} \:\forall i, j, \nonumber \\
& \sum_{j=1}^m a_{i,j} \leq 1 \:\forall i, \nonumber\\
\text{and}\quad  
& \sum_{i=1}^n a_{i,j} \leq 1 \:\forall j. \nonumber
\end{align}

While it is not difficult to solve (\ref{eq:1}) using well-established algorithms such as the Hungarian algorithm, the real challenge is how to consolidate an accurate estimate of the affinity score (i.e., $s_{i,j}$ is as close to the ground-truth as possible) due to the fact that we do not use annotated tracks.

\subsection{Overview of UnsMOT}

\begin{figure*}[t]
\centering
\includegraphics[scale=0.42]{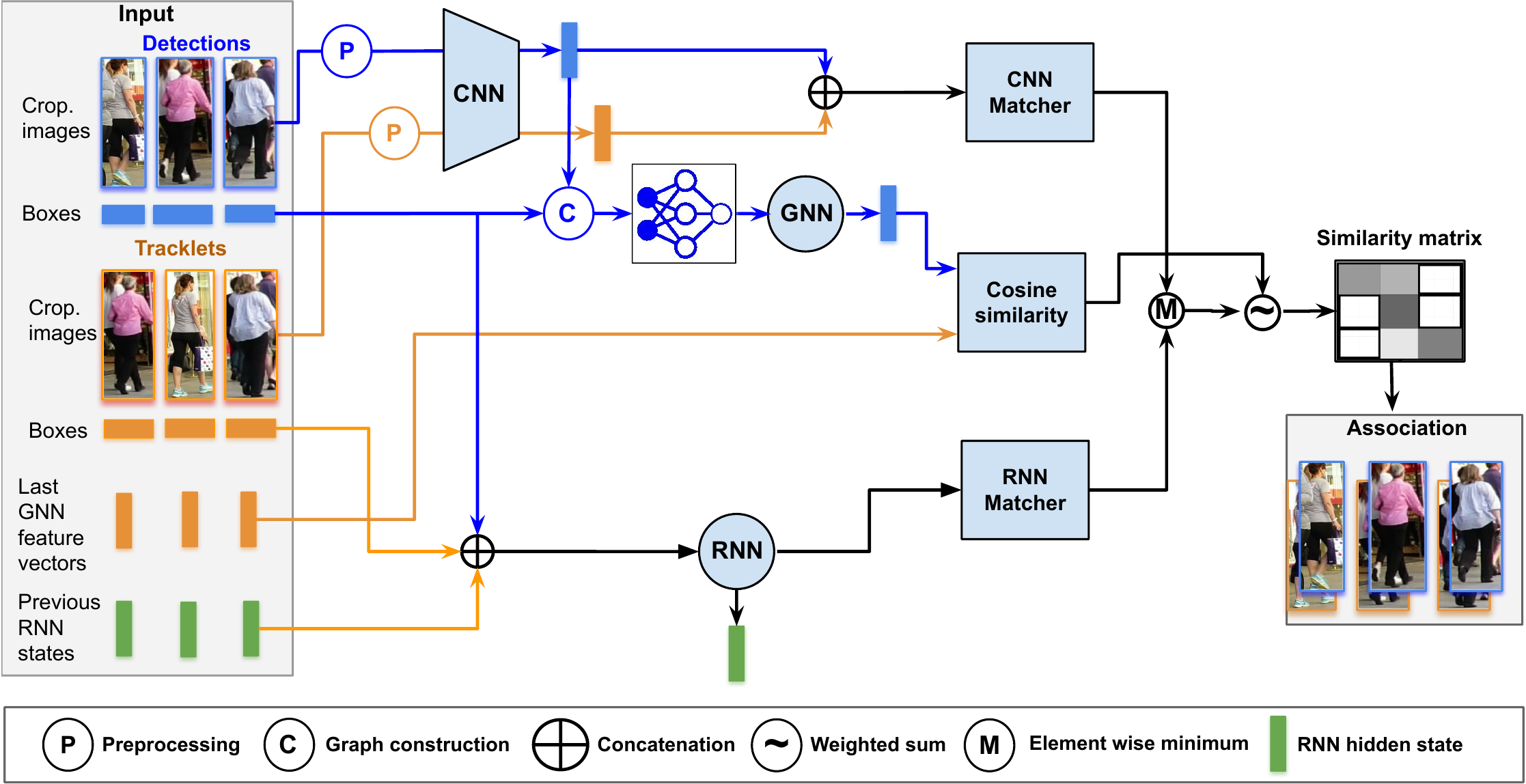}

\caption{\textbf{Schematic overview of our proposed framework.} Here, $n =  m = 3$.}
\label{fig:overview}
\end{figure*}

At each frame, we input the model with the set of detections at the current frame $\mathcal{D}$ and the set of active tracklets up to the current frame $\mathcal{T}$. As with other tracking-by-detection methods, we assume that the sets of detections $\mathcal{D}$ are provided.  As depicted in Fig. \ref{fig:overview}, our proposed framework consists of three branches, corresponding to three types of MOT object features, namely appearance, motion, and topology. This is different from previous unsupervised approaches which only focus on appearance and motion features. More specifically: 
\begin{itemize}
\item 1) In the first branch, after pre-processing, the cropped images of every detection $d_i \in \mathcal{D}$ and every tracklet $t_j \in \mathcal{T}$, denoted as $ img^ \mathcal{D} _i$ and $ img^\mathcal{T}_j$, respectively, are fed into a CNN model to obtain the appearance features. Then, we calculate a normalized pairwise appearance similarity matrix $S^A = \{s^A_{i,j}\}$ between extracted appearance features of every $d_i \in \mathcal{D}$ and $ t_j \in \mathcal{T}$ using a matcher network consisting of feed-forward layers. 
\item 2) In the second branch, we input the sets of bounding boxes $\{b_i^\mathcal{D}\}$,$\{b_j^\mathcal{T}\}$ and the previous RNN hidden state of the tracklets $t_j$ to an RNN model to calculate a normalized pairwise motion similarity matrix $S^M = \{s^M_{i,j}\}$ and an RNN hidden state representing the object motion between frames to be used in the next frame. 
\item 3) In the third branch, from $\mathcal{D}$, we construct a graph $G^\mathcal{D}(V^\mathcal{D},E^\mathcal{D},X^\mathcal{D})$, where a node $v_i^\mathcal{D} \in V^\mathcal{D} $ corresponds to a detection $d_i$. The node features of node $v_i^\mathcal{D}$, denoted as $x_i^\mathcal{D} \in X ^\mathcal{D}$, corresponds to the appearance feature of $d_i$ extracted from the CNN. Each pair of $v_i^\mathcal{D}$ and $v_j^\mathcal{D}$ corresponds to an entry $e_{i, j}^\mathcal{D}$ in the edge matrix $E^\mathcal{D}$, with $e_{i, j}^\mathcal{D} > 0$ if the Euclidean distance between the corresponding bounding boxes $b_i$ and $b_j$  is smaller than a threshold $T_{box}$, and the edge weight value $e_{i,j}^\mathcal{D}$ is obtained from a feed-forward layer. We then pass $G^\mathcal{D}(V^\mathcal{D},E^\mathcal{D},X^\mathcal{D})$ to our $L$-layers GNN to obtain the node embeddings matrix $H^{(\mathcal{D},l)}$ after every layer, with the $i$-th row $h_i^{(\mathcal{D},l)}$ as the node embedding vector of $d_i$ after the $l$-th layer. Intuitively,  $h_i^{(\mathcal{D},l)}$ captures the intra-frame topological structure of the $l$-hop neighborhood structure of the node $v_i$. We similarly denote $H^{(\mathcal{T},l)}$ as the node embedding matrix of active tracklets, with the $i$-th row $h_i^{(\mathcal{T}, l)}$ as the node embedding vector of $t_i$ at its last observed frame after the $l$-th layer. From $H^{(\mathcal{D},l)}$ and $H^{(\mathcal{T},l)}$,  we compute pairwise Cosine similarity between every pair of  $h_i^{(\mathcal{D},l)}$ and  $h_j^{(\mathcal{T},l)}$ to obtain the $l$-th layer intra-frame topological similarity score matrices $S^{(G,l)} = \{s^{G,l}_{i,j}\}$. 
\end{itemize}
The details of each branch will be further specified in later sections.

From the appearance, motion, and topological similarity matrices obtained above, we derive the final association matrix $S = \{s_{i,j}\}$ by first taking the element wise minimum of $S^A=\{s^A_{i,j}\}$ and $S^M = \{s_{i,j}^M \}$, then calculate its weighted sum with $S^{(G,l)} = \{s^{(G,l)}_{i,j}\}$ as follows:
\begin{equation}
\begin{split}
\label{eq:weighted_sum}
s_{i, j} = \alpha * \min(s^A_{i,j}, s^M_{i, j})   + \sum_{l = 1} ^ L \beta_l * s_{i,j}^{(G, l)},
\end{split}
 \end{equation} 
where $\alpha$ and $\beta_l$ are weighting hyper-parameters for each element subject to $\alpha + \sum_{l = 1} ^ L \beta_l = 1$.
Now, to solve (\ref{eq:1}), a simple greedy matching algorithm is applied to associate each detection $d_i$ to at most a single tracklet $t_j$, such that the affinity score $s_{i, j}$ between them is maximized.
\section{Proposed UnsMOT Framework } \label{sec:proposed_framework}
In this section, we describe the details of the CNN and RNN branches (Section \ref{sec:cnn_rnn}), then introduce our GNN-based proposal (Section \ref{sec:gcn}), and how to train all branches in an unsupervised learning manner (Section \ref{sec:training}).
\subsection{Appearance and Motion Similarity}\label{sec:cnn_rnn}
Appearance and motion similarity are vital parts of multi-object tracking. We extract appearance and motion similarity by employing a CNN and an RNN model similar to UNS20regress \cite{uns20}. For completeness, we only provide a summary, and the reader is advised to refer to the original paper for more details. 

Our CNN consists of six strided convolutional layers with a ReLU activation function after each layer except for the last layer, which outputs a 64-dimension feature vector encoding the object's appearance. Our RNN consists of four feed-forward layers that encode the object's motion up to the current frame to a 64 dimension vector.

To obtain the unnormalized appearance similarity matrix $\tilde{S}^A=\{\tilde{s}^A_{i,j}\}$, we pass the image crops $img^\mathcal{D}_i$ and $img_j^{\mathcal{T}}$to the CNN to obtain appearance features for the detection and tracklet in the current frame. We concatenate every pair of resulting feature vectors from detection $d_i \in \mathcal{D}$ with every resulting feature from tracklet $t_j \in \mathcal{T}$. Each pair is passed to a matching network consisting of 2 feed-forward layers, producing an unnormalized appearance similarity scores $ \tilde{S}^A=\{\tilde{s}^A_{i, j} \}$ with $\tilde{s}^A_{i, j}$ corresponding to the unnormalized appearance similarity scores between $d_i$ and $t_j$.

To obtain the unnormalized motion similarity matrix  $\tilde{S}^M=\{\tilde{s}^M_{i,j}\}$, we concatenate each pair of detection $b_i^\mathcal{D}$ and tracklet bounding box $b_j^\mathcal{T}$. Then we pass them to an RNN with two output heads. Both heads consist of a single feed-forward layer. The first head returns a motion similarity score $\tilde{s}^M_{i,j}$, while the second head outputs a hidden feature vector corresponding to the pair ($d_i$, $t_j$). Note that after finding the optimal association, for each tracklet $t_j$, we will keep the RNN feature vector corresponding to its pairing ($d_{i}$, $t_j$) as the tracklet's hidden feature to be used in inference in the next frames.

We normalize $\tilde{S}^A$ and $\tilde{S}^M$  using the same procedures in UNS20regress \cite{uns20}. Specifically, we define a normalize function $f(X)$ (change X to something else):
\begin{equation} \begin{split}
\begin{split}
\label{eq:softmax_nomalizing}
f(X) = \min (\text{softmax}_{\text{col}}(X), 
\text{softmax}_{\text{row}}(X))
\end{split}
\end{split} \end{equation} 
with $\min$ denoting element-wise minimum, $\text{softmax}_{\text{row}}(X)$ and $\text{softmax}_{\text{col}}(X)$ denoting the softmax function applied on the matrix $X$ row-wise and column-wise, respectively. Then, we normalize  $\tilde{S}^A $ and $\tilde{S}^M $ as follows:
\begin{equation} \begin{split}
\begin{split}
S^A = f(\tilde{S}^A), 
S^M = f(\tilde{S}^M).
\end{split}
\end{split} \end{equation} 
\subsection{Intra-frame Topological-based Association Using Graph Neural Network}\label{sec:gcn}
\begin{figure*}[t]
 \centering
\includegraphics[width=\textwidth]{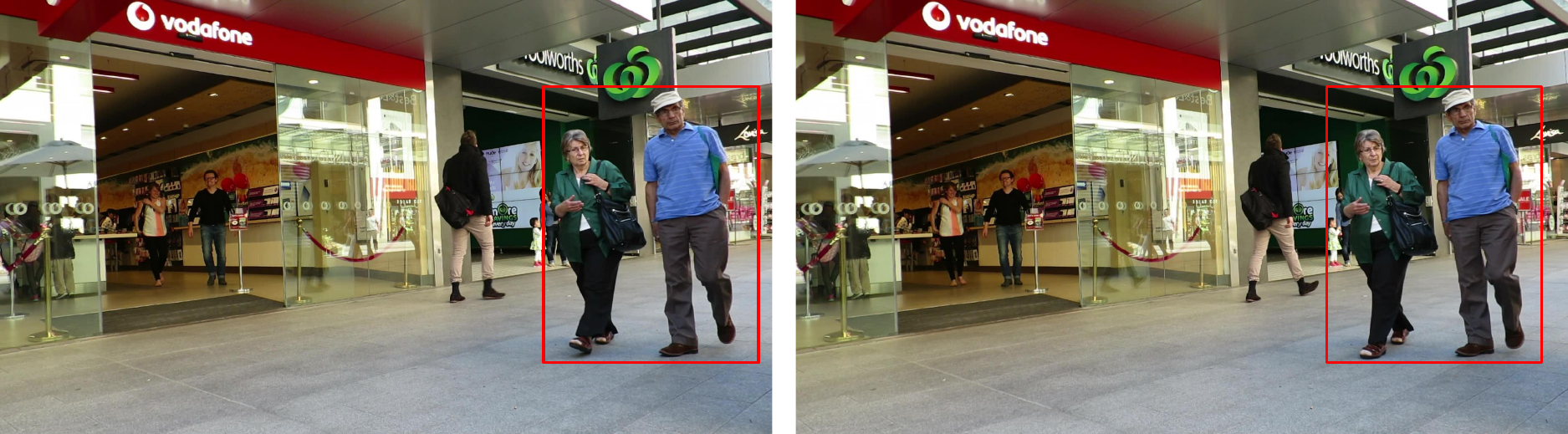}
\captionsetup{justification=centering}
\caption{An example demonstrating that objects that appeared near each other (highlighted in the red box) in a frame are likely to maintain their relative position in the next frame)}
 \label{fig:observation}
\end{figure*}
While appearance and motion features are generally sufficient for robust tracking, they tend to suffer from lower performance on sequences with moving cameras and occluded objects. To alleviate this problem, we propose a novel GNN-based approach, based on a simple observation: objects that appeared near each other in a previous frame are likely to maintain their relative position in the next frame. Fig. \ref{fig:observation}
provides an example illustrating our observation. In the previous frame, we observe that a person in a pink shirt is walking near a person in a blue shirt and a person in a red shirt. In the next frame, it makes sense to assume that the same person is more likely to be walking near people in blue shirts and red shirts. Our framework exploits this assumption to employ the intra-frame topology between objects as a strong clue to adjust the detection-tracklet association. This intra-frame topology information can be represented by a graph, as we will show in Section \ref{sec:graphconstruction}, leading us to design a GNN-based network branch for computing intra-frame topology-based association. This topological feature is naturally robust to camera movements as well as changes in the object position and visibility, which heavily degrades an object's features, by augmenting it with additional cues from the neighboring objects. 
\subsubsection{Graph Construction} \label{sec:graphconstruction}
To enable the use of a GNN model, we devise a simple graph construction module. The module inputs $b^{\mathcal{D}}$ and their corresponding appearance feature vectors and outputs a graph $G^\mathcal{D}(V^\mathcal{D},E^\mathcal{D},X^\mathcal{D})$ consisting of nodes \(v_i \in V\) corresponding to every detections $d_i \in \mathcal{D}$, edge matrix \(E^\mathcal{D}\) with $e_{i, j}^\mathcal{D} \neq 0$ if node $v_i$ and $v_j$ are connected, and a feature vector matrix \(X^\mathcal{D}\), with \(x_i^\mathcal{D}\) as the appearance feature vector extracted by our CNN for node \(v_i^\mathcal{D}\).

We consider a few different approaches to construct the edge matrix \(E^\mathcal{D}\). A naïve strategy is to build a fully-connected graph where each node connects to every other node. However, a graph constructed in this manner does not provide any intra-frame topological information, therefore not useful. Another approach is connecting a node to \(k\) other nodes with the smallest geometrical distance. The resulting graph would naturally contain a topological structure, which can then be exploited using our method. In practice, we find that the graph's topology changes too frequently and drastically, reducing the effectiveness of our model. We instead propose to connect a pair of nodes $v_i^\mathcal{D}$ and $v_j^\mathcal{D}$ in the graph if their normalized bounding boxes have a Euclidean distance $dist(b_i^\mathcal{D}, b_j^\mathcal{D})$ smaller than a threshold \(T_{box} \). The resulting graph has a natural topology between objects at different distances while not changing as frequently and drastically, thus providing a more stable association. 

We also calculate a weight for each edge $e_{i, j}^\mathcal{D}$ following the procedures in \cite{gcnnmatch}. We begin by concatenating $x_i^\mathcal{D}$ with $x_j^\mathcal{D}$, and bounding box $b_i^\mathcal{D}$ with $b_j$. Then, we concatenate the two resulting vectors and pass them through a feed-forward layer that outputs a single scalar  value to represent our edge weights.

\subsubsection{Embedding of Geometric Information}
 To extract the geometric information between objects, we select the GCN formulation as described in \cite{gcn}  through empirical means. Specifically, the $l$-th order geometric embedding $H^{(\mathcal{D}, l)}$ of a graph $G^\mathcal{D}$ is calculated as follows:

\begin{equation} \begin{split}
\label{eq:gcn}
H^{(\mathcal{D}, l)} = {({\hat{D}})}^{-\frac{1}{2}} \hat{E} {({\hat{D}})}^{-\frac{1}{2}} W^l X^\mathcal{D},
\end{split} \end{equation} 

\noindent where $H^{(\mathcal{D}, l)}$ is the hidden feature after $l$ GCN layers, with the $i$-th row $h_i^{(\mathcal{D},l)}$ as the node features of detection $d_i$ after $l$ layers, $\hat{E} = E^\mathcal{D} + I$ with $I$ is the identity matrix, $\hat{D}=\deg(\dot{E} + I)$ with $\dot{E} = \{\dot{e}_{i, j}\}$ and $\dot{e}_{i,j} = 1$  if  $e_{i, j} \neq 0$, else $\dot{e}_{i, j} = 0$,  and $W^l$ as the weight matrix of the $l$-th GCN layer.
By propagating and aggregating features of each node's neighborhood to itself, after \(l\) layers, each node's feature vector now contains information about all other nodes in its \(l\)-hop neighborhood. Deeper layers contain more global but less distinctive topological features, while shallower layers contain topological features that are more distinctive  but sensitive to changes in the neighborhood structure between frames. 

\subsubsection{Topological Similarity Matrix}
We now describe the final component of our framework, the intra-frame topological similarity matrix. We begin by constructing a graph from the detections bounding box and images, and passing the resulting graph through a GNN to obtain node features $h_i^{(\mathcal{D},l)}$ of every detection $d_i$ after every layer $l$. In our study, we propose to calculate the topological similarity matrix using the node feature vector after every layer of the GNN instead of only the topological similarity at the final layer. Formally, we calculate the Cosine similarity for detection $d_i$ and the tracklet $t_j$ after the $l$-th layer of the GNN as:
\begin{equation} \begin{split}
\begin{split}
\cos (h^{(\mathcal{D},l)}_{i}, h_{j}^{(\mathcal{T},l)}) =\frac {{h^{(\mathcal{T},l)}_i \cdot h_j^{(\mathcal{D},l)}}}  {\|h^{(\mathcal{T},l)}_{i}\| \|h_{j}^{(\mathcal{D},l)} \|}
\end{split}
\end{split} \end{equation} 
\noindent with $h_j^{(\mathcal{T},l)}$ as the node feature vector of tracklet $t_j$ computed at its last observed  frame.

\subsection{Training Procedure}\label{sec:training}

Our model is trained in two stages. We first train the CNN, RNN, and their matching models following the method described in \cite{uns20}. Then, we train our GNN based on features obtained from the trained CNN model using an unsupervised loss function.

\subsubsection{CNN and RNN}
 For completeness, we provide a summary of the training process for CNN and RNN models, and the reader is recommended to refer to the original paper for more details. We employ the visual-spatial hiding scheme of \cite{uns20}. Specifically, we train on samples of $N$ frames from the training set. For the CNN branch, only the first and last frame of the sample is used. Each branch only calculates the output for objects that appeared in the first frame of the sample. We apply the CNN and RNN branches using the procedures described in Section \ref{sec:cnn_rnn} on the sample to obtain two outputs $O_1=\{o_{1,i, j}\}$ and $O_2=\{o_{2,i, j} \}$ corresponding to each branch. The authors then proposed a novel loss that, when optimized, enforces consistency between these outputs, thus forcing the CNN and RNN to learn relevant appearance and motion features. Formally:
\begin{equation} \begin{split}
J_1 = -\sum_i \log \sum_j (o_{1,i,j} {o}_{2,i,j} c_{i,j})
\end{split} \end{equation} 
with $C = \{ c_{i,j} \}$ as a binary matrix added to prevent the models from converging at a local optima, and $c _{i,j} = 0$ if objects $i$ and $j$ are unlikely to correspond to the same objects, which is determined by a floodfill-like algorithm.
\subsubsection{GNN}
\begin{figure*}[t]
\centering
\includegraphics[width=\textwidth]{
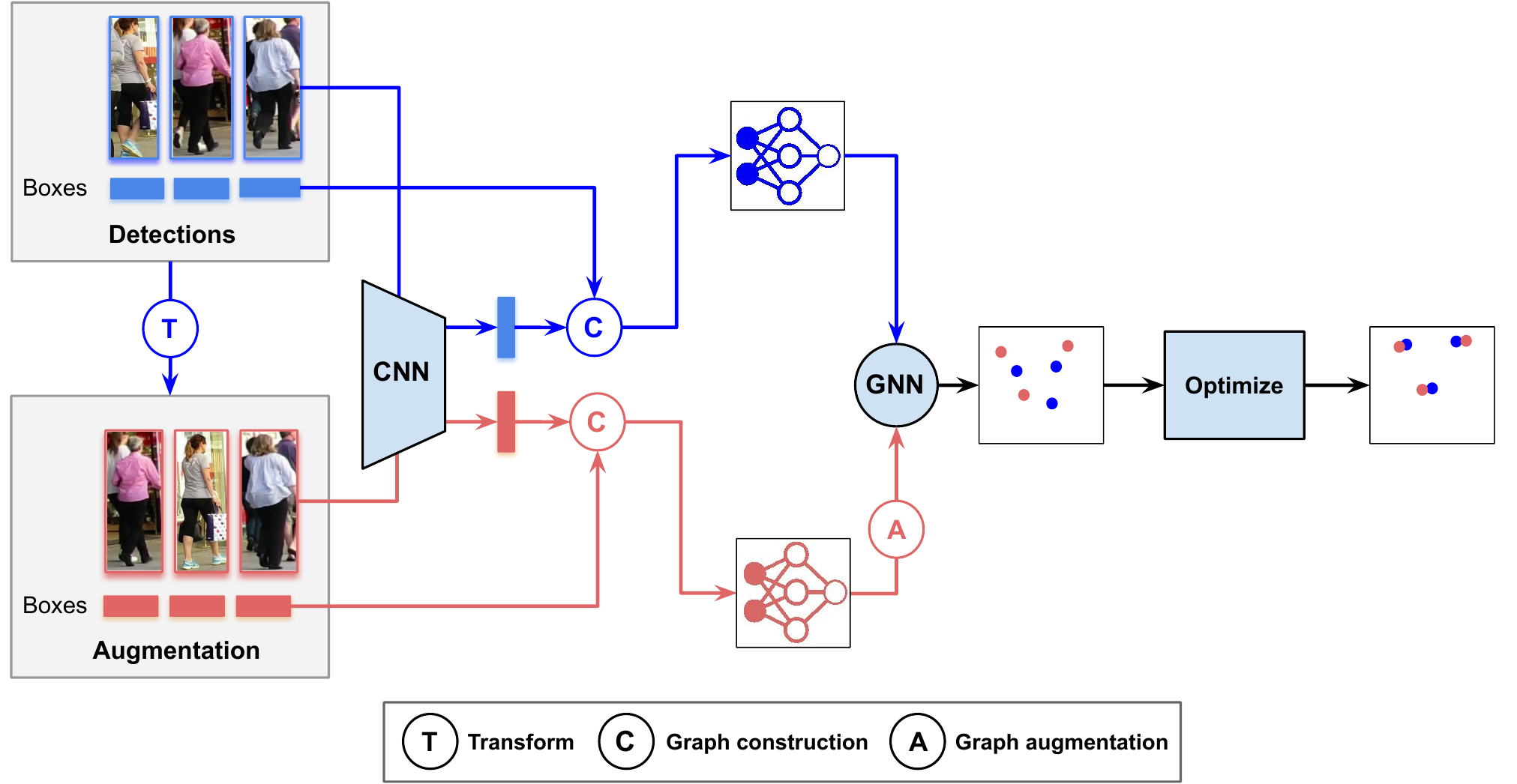
}
\caption{Our proposed unsupervised GNN training workflow.}
\label{fig:gcn_train}
\end{figure*}

The second stage of our training procedure is to learn the GNN model in an unsupervised manner. To this end, we train our model using the layer-wise reconstruction loss described in \cite{gradalign} and the noise adaptivity loss in \cite{galign}. Fig \ref{fig:gcn_train} describes our training procedure. In particular, the layer-wise reconstruction loss is defined as:
\begin{equation} 
\begin{split}
\label{eq:reconstruction_loss}
    J_{r}(G^\mathcal{D}) = \| ({\tilde{D}^{l}})^{-\frac{1}{2}} \tilde{E}^{l} ({\tilde{D}^{l}})^{-\frac{1}{2}} - H^{(\mathcal{D},l)} ({H^{(\mathcal{D},l)}})^T \|
\end{split} 
\end{equation} 
with $({H^{(\mathcal{D},l)}})^T$ denoting the transpose matrix of ${H^{(\mathcal{D},l)}}$, and $\tilde D $ as a diagonal matrix containing the sum of the $l$ first powers of $\hat{E}$, or more formally $\tilde{E}^{(l)} =\sum_{k=1}^l {({\hat{E}})}^{k}$, $\tilde{D}_{i, i}^{l} =  \sum_{j} {{\tilde{E}_{i, j}}}^{l}$.   
This loss function encourages nodes with similar neighborhood topologies to have embeddings with higher Cosine similarity, and nodes with different neighborhood topologies to have embeddings with lower Cosine similarity.

Recall that our final goal is to align the tracklets and detections graphs. As objects change between frames, so do the constructed graphs. The above loss alone would fail to account for these changes. Following GAlign \cite{galign}, we also enforce our model to adapt to such changes by adding an adaptivity loss. The loss minimizes the difference between resulting feature vectors obtained from applying the GNN model on the original graph $G^\mathcal{D}(V^\mathcal{D},E^\mathcal{D},X^\mathcal{D})$ and a graph $\bar{G}^\mathcal{D}(\bar{V}^\mathcal{D}, \bar{E}^\mathcal{D}, \bar{X}^\mathcal{D})$ augmented from $G^\mathcal{D}(V^\mathcal{D},E^\mathcal{D},X^\mathcal{D})$:
\begin{equation}
\begin{split}
J_{a}(G^\mathcal{D}, \bar{G}^\mathcal{D}) = \sum _{i, j} \sum_l \sigma (\| (h_i^\mathcal{D})^l - 	({\bar{h}^\mathcal{D}}_j)^l \|)
\end{split} 
\end{equation} 
with $\bar{h_{j}}^\mathcal{D}$ as the node hidden feature of augmented node $\bar{v_{j}}^\mathcal{D} \in  \bar{V}^\mathcal{D}$. To obtain $\bar{G}^\mathcal{D}$, the authors in \cite{galign} intentionally added noises to $G^\mathcal{D}$ by adding/removing edges and attribute noise by randomly adjusting values  in each feature vector. We adapt these operations to align better to the \prob{} problem. Specifically, based on the original proposal, we propose three following augmentation operations to obtain the augmented graph $\bar{G}^\mathcal{D}(\bar{V}^\mathcal{D}, \bar{E}^\mathcal{D}, \bar{X}^\mathcal{D})$  augment $G^\mathcal{D}(V^\mathcal{D},E^\mathcal{D},X^\mathcal{D})$:
(1) Adding/removing random edges,
(2) Removing random nodes,
(3) Changing initial node features.

We now go into the details of each operation. Operation (1) randomly removes/adds edges from/to the graph with a probability $p_1$. Operation (2) remove nodes from the graph with a probability $p_2$. For operation (3), since our initial node feature vector $x_i$ is extracted from a CNN, instead of directly changing each node feature vector, it would make more sense to apply augmentation on the initial image crop before applying the CNN. To this end, we apply commonly used image augmentation techniques on both the cropped image and bounding box with a probability $p_3$ like rotating, translating, and more before passing them to our CNN. We also control their strength with a single hyper-parameter $\epsilon$ that linearly scales to each augmentation operation magnitude.

Finally, we combine $J_r$ and $J_a$ to get the final training loss for the GNN:
\begin{equation} 
\begin{split}
J_2 = \gamma J_r(G^\mathcal{D}) + (1 - \gamma) J_a(G^\mathcal{D}, {\bar{G}^\mathcal{D}})
\end{split} 
\end{equation} 
with $\gamma$ as a weighting hyper-parameter.
The above loss does not require labeled data. As such, it is a natural extension of the self-supervised model in \cite{uns20}.

\section{Experiments}\label{sec:infer}
\subsection{Datasets}
Recently, there have been many datasets focusing on the MOT problem \cite{kitti, MOT16}. We evaluate our framework on the MOT challenge \cite{MOT16}. They are a collection of public, academic datasets commonly used to evaluate tracking method performance. For MOT, we examine our framework for the MOT16, MOT17, and MOT20 versions. MOT16 contains 14 video sequences divided equally into a train and test set, filmed with static and moving cameras. MOT17 includes identical video sequences as MOT16 but provides more accurate ground truth and two additional public detections generated from a Faster-RCNN \cite{ren2015faster} and SDP \cite{sdp}. MOT20 contains 8 video sequences equally divided into a training and testing set, filmed in different challenging environments. We split all dataset training sets into parts for training, validation, and testing. Specifically, for each video, we define a fixed percentage of frames for each part. For MOT, with each video, we split 80\% of the frames into the training set and the remaining 20\% for validation. Note that video annotation is not utilized for training the model in our proposed UnsMOT method.
\subsection{Evaluation Metrics}
 As our proposed framework is strictly an association method, we chose to focus on 2 comprehensive metrics: IDF1 and HOTA. In short, these metrics measure the performance of a tracker by measuring the accuracy of the predicted tracks versus the ground truth tracks. IDF1 \cite{IDF1} focuses on measuring association performance, while HOTA \cite{hota} is a newly proposed metric that aims to balance measuring detections and association. MOTA \cite{MOTA} is another widely used metric in \prob{} literature but heavily favors measuring detection accuracy \cite{hota}, thus is only provided for completeness.   
\subsection{Experimental Settings}
We train our CNN, RNN, and matcher models using the same datasets used in \cite{uns20}, which includes five hours of  YouTube walking tours videos, and every video from the PathTrack dataset \cite{pathtrack}. Our best-performing model hyper-parameters were found using the grid search strategy. Our final model had 2 GCN layers, with the weighting hyper-parameters $\alpha = 0.7$, $\beta_1 = 0.2$, and $\beta_2 = 0.1$. We trained the models with an Adam optimizer with the learning rate set to $2e-6$. The threshold $T_{box}$ is set to $0.1$.  The augmentation strength $\epsilon$ is set to $0.01$. The operation probabilities $p_1$, $p_2$ and $p_3$ are set to $0.1$.  The loss weight $\gamma$ is set to $0.8$. Other hyper-parameters are kept the same as in the original paper. The implementation of UnsMOT is conducted using PyTorch \cite{pytorch} and PyTorch-Geometric \cite{pygeo}. All experiments are conducted on a server with Tesla M10 GPU, Intel(R) Xeon(R) Silver 4110 CPU, and 32GB of system RAM. The source code is included in the supplemental and will be publicly released in our MOT challenge leader board entry upon acceptance of the paper.

At inference, to robustify the final association score $s_{i, j}$, for the CNN and GNN branch, we match 5 additional past observations of each tracklets, and average their association scores. Similar to recent methods, we also preprocess the public detections using the same method in Tracktor++ \cite{tracktor}.
\subsubsection{Baselines}
We compare our approach to the state-of-the-art unsupervised and  similar supervised methods, including UNS20regress \cite{uns20},         SORT  \cite{sort},           V-IOU \cite{iou}, GCNNMatch \cite{gcnnmatch}, 
GSM-Tracktor \cite{gsmtrack}, 
 Tracktor++v2 \cite{tracktor},
 DeepMOT \cite{xu2020train}, CenterTrack \cite{centertrack}, and 
 LSST\cite{feng2019multi}. While there exist supervised methods that perform better than ours on the MOT datasets, which typically use larger feature extractors and are trained using labeled data, we have chosen to only include baseline supervised methods based on both popularity and similarity in terms of approaches to ours to highlight our method promising performance,  as the comparison between our unsupervised method and supervised methods is not the main focus of this paper.

\subsection{Experimental Results}
 In this section, our empirical study is designed to answer the following questions:
 \newcommand\litem[1]{\item{\bfseries #1}}
 \begin{enumerate}
 \item [Q1:] How much does UnsMOT improve from the state-of-the-art unsupervised method?
 \item [Q2:] How does UnsMOT improve tracking performance? 
 \item [Q3:] How does each component of our proposed model contribute to the final performance?
 \item [Q4:] How expensive is the proposed method's computational complexity in comparison with other multi-object tracking methods?
\end{enumerate}
 
 \subsubsection{Comparison With State-of-the-art Methods (Q1)}
We trained our models on the entire MOT16, MOT17, and MOT20 training sets and submitted UnsMOT to the MOT challenge leader boards. Table \ref{result} shows the performance comparison of UnsMOT and competitive schemes on the MOT datasets. Since the results are taken from the MOT Challenge website,\footnote{\url{https://motchallenge.net/}}, we omit methods without publicly available results for each dataset. 

Overall, UnsMOT consistently outperforms all but one competitive method on the datasets in terms of IDF1 and HOTA. Notably, UnsMOT achieves state-of-the-art performance for the unsupervised MOT problem by outperforming the most recent method, termed UNS20regress. It is also interesting to see the superiority of UnsMOT to popular supervised techniques despite our proposed framework not requiring any expensive track-level information. 
\begin{table*}[ht]
\centering
 \caption{Comparison of our proposed framework with state-of-the-art online supervised and unsupervised trackers on the public detections MOT Challenge Benchmark. Here, the best and second-best methods for each case are highlighted using \textbf{bold} and \underline{underline}, respectively.}
\setlength{\tabcolsep}{2mm}
 \begin{tabular}{c c c c c c c c c c c c c} 
 \Xhline{3\arrayrulewidth}
  \multicolumn{12}{c}{MOT 2016} \\
  \Xhline{2\arrayrulewidth}
  Method & Unsup & HOTA$\uparrow$ &  MOTA$\uparrow$ & IDF1$\uparrow$ & MT & ML & FP & FN & IDSw & Frag & Inference time (s)\\ [0.5ex]
 \Xhline{2\arrayrulewidth}
 UnsMOT(Ours) & \cmark & \textbf{48.0} & \textbf{57.6} & \textbf{60.6} & 190  & 266  & 6,738 & 73,214 & 439  & 652 & 0.02 \\ 
 GCNNMatch \cite{gcnnmatch} & \xmark & 44.6 & \underline{57.2} & 55.0 & 174  & 258  & 3,905 & 73,493 & 559  & 847 & 2.51\\ 
 GSM-Tracktor\cite{gsmtrack} &  \xmark & \underline{45.9} & 57.0 & \underline{58.2} & 167  & 262  & 4,332 & 73,573 & 475  & 859 & - \\
 Tracktor++v2\cite{tracktor} & \xmark & 44.6 & 56.2 & 54.9 & 157  & 272  & 2,394 & 76,844 & 617  & 1,068  & 0.97 \\
 DeepMOT\cite{xu2020train} & \xmark & 42.2 & 54.8 &53.4 & 145  & 281  & 2,955 & 78,765 & 645  & 1,515 & - \\
 \Xhline{3\arrayrulewidth}
  \multicolumn{12}{c}{MOT 2017} \\
 \Xhline{2\arrayrulewidth}
  Method & Unsup & HOTA$\uparrow$ &  MOTA$\uparrow$ & IDF1$\uparrow$ & MT & ML & FP & FN & IDSw & Frag & Inference time \\ [0.5ex]
 \Xhline{2\arrayrulewidth}
 UnsMOT(Ours) & \cmark & \underline{48.0} & \underline{57.6} & \textbf{60.6} & 622  & 820  & 16,197 & 221,769 & 1,359  & 2,191 & 0.02  \\ 
 UNS20regress \cite{uns20} & \cmark & 46.4 & 56.8 & \underline{58.3} & 538  & 880  & 11,567 & 230,645 & 1,320  & 2,061 & 0.09 \\
 SORT \cite{sort} & \cmark & 34.0 & 43.1 & 39.8 & 295  & 997  & 28,398 & 287,582 & 4,852  & 7,127 & 0.0004 \\
 V-IOU\cite{iou} & \cmark & 33.5 & 45.5 & 39.4 & 369  & 953  & 19,993 & 281,643 & 5,988  & 7,404 & 0.0005\\
 GCNNMatch \cite{gcnnmatch} & \xmark & 45.4 & \underline{57.3} & 56.3 & 575  & 787  & 14,100 & 225,042 & 1,911  & 2,837 & 2.51 \\
GSM-Tracktor\cite{gsmtrack} & \xmark & 45.7 & 56.4 & 57.8 & 523  & 813  & 14,379 & 230,174 & 1,485  & 2,763  & - \\
CenterTrack \cite{centertrack} &  \xmark & \textbf{48.2} & \textbf{61.5} & 59.6 & 621  & 752  & 14,076 & 200,672 & 2,583  & 4,965 & - \\
Tracktor++v2\cite{tracktor} & \xmark & 44.8 & 56.3& 55.1 & 498  & 831  & 8,866 & 235,449 & 1,987  & 3,763 & 0.97   \\
DeepMOT\cite{xu2020train} & \xmark & 42.4 & 53.7 &53.8 & 458  & 861  & 11,731 & 247,447 & 1,947  & 4,792 & -  \\
LSST\cite{feng2019multi} & \xmark & 44.3 & 52.7 & 57.9 & 421  & 863  & 22,512 & 241,936 & 2,167  & 7,443   & - \\
  \Xhline{3\arrayrulewidth}
  \multicolumn{12}{c}{MOT 2020} \\
   \Xhline{2\arrayrulewidth}
 Method & Unsup & HOTA$\uparrow$ &  MOTA$\uparrow$ & IDF1$\uparrow$ & MT & ML & FP & FN & IDSw & Frag & Inference time \\ [0.5ex]
 \Xhline{2\arrayrulewidth}
 
UnsMOT(Ours) & \cmark & \textbf{42.7} & \textbf{54.7} & \textbf{52.4} & 409  & 302  & 10,731 & 221,649 & 1,829  & 1,857 & 0.02 \\ 
UnsupTrack \cite{simpleuns} & \cmark & 41.7 & 53.6 & 50.6 & 376  & 311  & 6,439 & 231,298 & 2,178  & 4,335 & - \\
SORT \cite{sort} & \cmark & 36.1 & 42.7 & 45.1 & 208  & 326  & 27,521 & 264,694 & 4,470  & 17,798  & 0.0004\\
GCNNMatch \cite{gcnnmatch} & \xmark & 40.2 &  \underline{54.5} & 49.0 & 407  & 317  & 9,522 & 223,611 & 2,038  & 2,456 &  2.51\\
Tracktor++v2\cite{tracktor} & \xmark &  \underline{42.1} & 52.6 &  \underline{52.7} &  365  & 331  & 6,930 & 236,680 & 1,648  & 4,374 & 0.97 \\
 \Xhline{3\arrayrulewidth}   
 \end{tabular}
\label{table:experimental_results}
 \label{result}
\end{table*}


\noindent\textbf{Comparative Study Among Unsupervised Methods.}
Out of all the unsupervised methods, traditional methods like SORT and V-IOU perform the worst, with significantly lower IDF1 and HOTA scores than deep-learning-based approaches. Their poor performance can be attributed to the fact that SORT and V-IOU rely on traditional hand-crafted features and thus are less robust. Specifically, SORT uses the Kalman filter to associate tracks based on estimated motion, while V-IOU instead tracks by combining IOU threshold and hand-crafted visual features. Compared with unsupervised learning approaches that only rely on appearance features like UnsupTrack on the MOT17 dataset, our proposed model has 1\% higher HOTA and nearly 2\% higher IDF1. Compared with approaches that rely on a combination of appearance and motion features like UNS20regress on the MOT20 dataset, our model has 2\% higher HOTA and 2\% higher IDF1. This suggests that our addition of topological features along with appearance and motion features results in more consistent tracks across time. 

\noindent\textbf{Comparative Study Against Supervised Methods.}
Surprisingly, compared to recent supervised methods, our model consistently outperforms them on almost every metric, despite not using any expensive video-level labels. Compared to methods like CenterTrack and Tracktor++v2, which exploits object detection models to also predict bounding box offsets from the current frame to the next frame, our proposed framework outperforms them both in terms of IDF1, only losing to CenterTrack by 0.2\% HOTA. This can be attributed to CenterTrack's more accurate bounding boxes than the boxes used in our method, as evident by their significantly higher MOTA score. It is worth pointing out that although the supervised CenterTrack model outperformed us in terms of HOTA and MOTA, our proposed model is still able to outperform the unsupervised CenterTrack model in terms of the IDF1 metric. 

\noindent\textbf{Comparative Study Among GNN-based Methods.} Our proposed method also consistently surpasses the previous graph-based supervised methods like GCNNMatch and GSM-Tracktor. Specifically, GCNNMatch aims to utilize a GNN to enrich the features of different objects in two adjacent frames with features from other objects within a certain proximity, while GSM-Tracktor aims to calculate the similarity between the object's constructed graph in different frames. This key difference robustifies our model against camera movement and partial occlusion and enables model training with an unsupervised graph matching loss. In addition, these methods only use CNN and GCN, while our proposed model also exploits the object's motion using an RNN. 
\subsubsection{Qualitative Studies (Q2)} \label{sec:qualitative}
\begin{figure*}[h]
\centering
\includegraphics[scale=0.4]{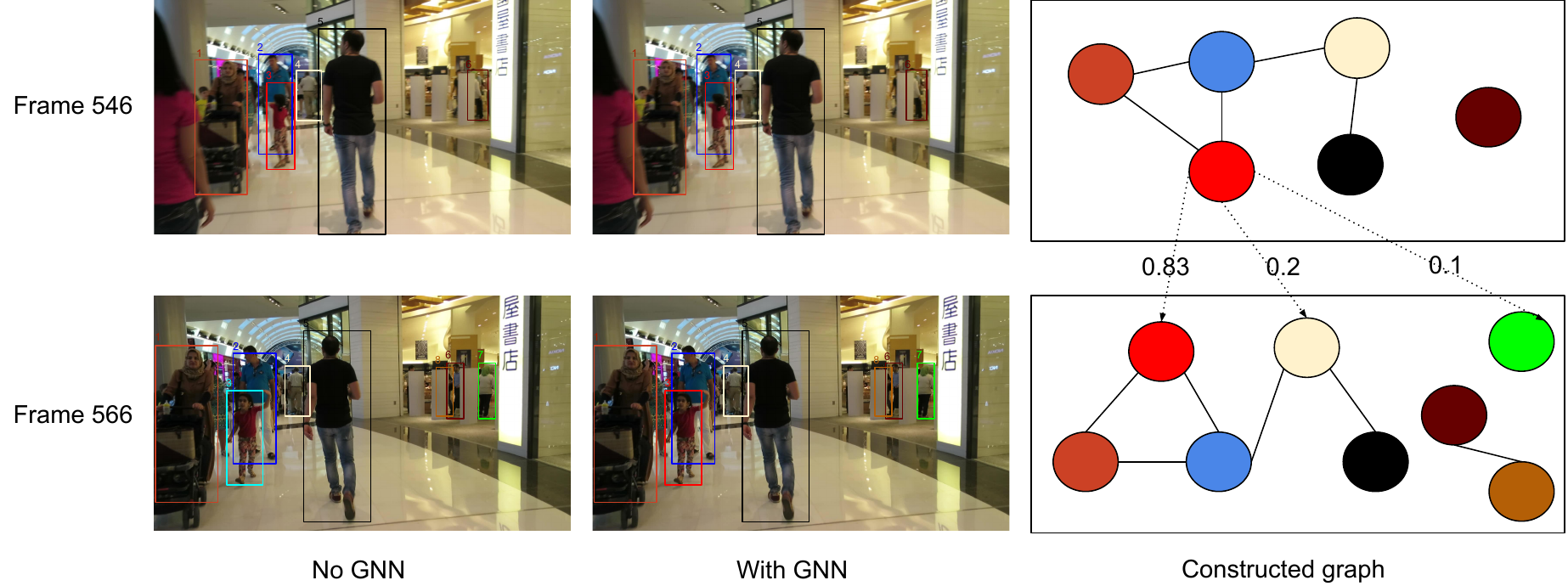}
\includegraphics[scale=0.4]{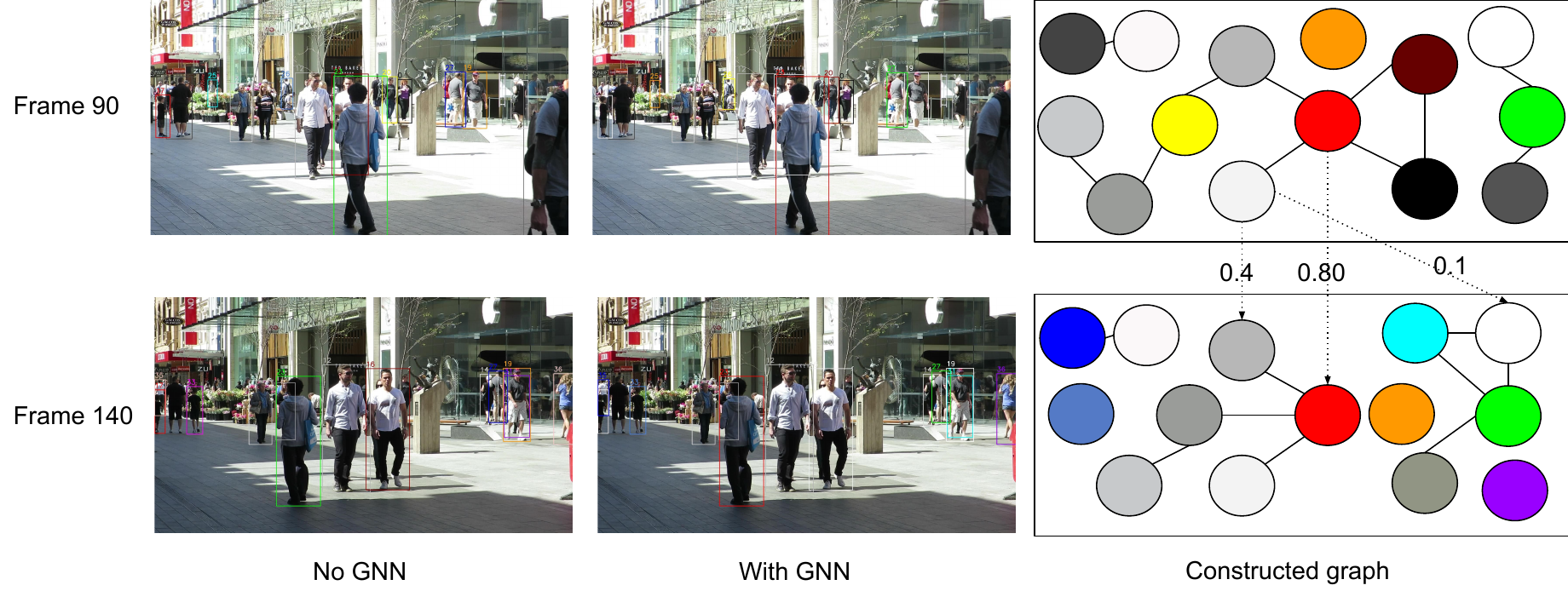}
\captionsetup{justification=centering}
\caption{Qualitative studies of our method on the MOT17-11 test sequence. Our method consistently tracks objects even with  occlusions and large bounding box overlaps}
\label{fig:qualitative}
\end{figure*}

We show two qualitative results in Fig. \ref{fig:qualitative}. Our method robustly tracks objects even with moving cameras and heavy occlusions. For both cases, we show that our method can consistently track objects in large groups, even though their bounding boxes highly overlap. Fig. \ref{fig:qualitative} illustrates two case studies of the information captured by our GNN module. We depict subjects with the same predicted identity (ID) with the same bounding box color. With no GNN present, an ID switch occurred for the subject colored in bright red in the first case, and for the subject in white in the second case. Specifically, the ID switch occurred because of low appearance and motion similarity between the same object across frames. In both cases, these low scores are caused by occlusion or moving cameras. With GNN present, no ID switch happened. This is due to the additional topological information captured by the GNN. In both cases, the topological similarity of the same object across frames remains high due to similar neighborhoods even when the appearance and motion similarity are low, and thus the object is still able to be consistently tracked. 

\subsubsection{Ablation Studies (Q3)} \label{sec:ablation}
To better understand the influence of different proposed components on the final performance, we perform several experiments by dividing sequences of  MOT17 with the F-RCNN object detector into two sets, with the first 80\% of each video used as training data and the remaining 20\% for validation. 

\textbf{Role of CNN and RNN}.
We first evaluate the role of the CNN and RNN models in the simplified systems without the GNN. We report the tracking results when having both CNN and RNN, CNN only, and RNN only in the first three rows in Table \ref{table:ablation}. As can be seen, our system obtains the best scores when having both models. The CNN-only system performs better than the RNN-only one, which is reasonable since appearance is the most obvious feature for tracking.

\textbf{Effects of Node Features}.
In this line of ablation study, we show the importance of the node features to our GNN branch. We decoupled the node features from the GNN branch by removing the node features from Equation. \ref{eq:gcn}. All other hyper-parameters are kept the same as our best performing model described in Section \ref{sec:infer}. We report the model's performance as "No node features" in Table. \ref{table:ablation}. As evident from the significant performance decrease when compared with models utilizing node features, the node features plays an important role in the final model performance.

\textbf{Effects of graph construction}.
In Section \ref{sec:graphconstruction}, we discussed possible graph construction techniques. In this ablation study line, we investigate the performance of an alternative graph construction strategy. Specifically, we modify the graph construction algorithm in \ref{sec:graphconstruction} to instead connect a node to \(k\) other nodes with the smallest geometrical distance, referred to as UnsMOT-$k$. No additional changes to our framework were made. The performance of this model is shown in Table \ref{table:ablation}. As evident from the final model's performance, although the addition of GNN layers is still beneficial to the performance, this graph construction technique yields lower HOTA and IDF1 scores compared to the strategy described in \ref{sec:graphconstruction}.

\textbf{Effects of GNN Layers}.
It is common practice to use only a couple of GNN layers. We ablate the number of GNN layers by training and evaluating models with the number of layers increasing from 0 to 3. The tracking results are reported in Table \ref{table:ablation}. When incorporating GNN, even with only one layer, both HOTA and IDF1 scores increase by around 1\%, proving the significance of modeling the topological features. When increasing the number of GNN layers from 1 to 2, the tracking performance still improves, but the gains are less noticeable, with 0.37\% for HOTA and 0.80\% for IDF1. The network performance peaked at 2 GNN layers, and when we increased the number of layers to 3, both metrics started decreasing, suggesting that the deeper layers hurt the system performance by boosting the affinity score of connected nodes in larger regions instead of focusing on a small region. Based on this ablation study, we use 2-layer GNN in our system with all other experiments.
\begin{table}[t]
\centering

 \caption{Ablation study of the effects of the number of layers on the tracking performance on the validation set. Here, the best and second-best methods for each case are highlighted using \textbf{bold} and \underline{underline}, respectively. All other hyper-parameters are kept  unchanged.}
\setlength{\tabcolsep}{4mm}
 \begin{tabular}{c c c } 
 \Xhline{2\arrayrulewidth}
  \ Model name   & HOTA$\uparrow$ & IDF1$\uparrow$ \\ [0.5ex]
 \Xhline{2\arrayrulewidth}
   No GNN & 58.59  & 64.82 \\ 
   No RNN and GNN & 58.22 & 64.63 \\
   No CNN and GNN & 55.97 & 60.25 \\
   No node features & 58.75 & 65.19 \\
  CNN, RNN and 1 layer GNN & \underline{59.23}  & \underline{65.59} \\ 
   CNN, RNN and 2 layer GNN   & \textbf{59.60} & \textbf{66.39} \\ 
 CNN, RNN and 3 layer GNN   & 59.57 & 66.26  \\ 
    UnsMOT-3 & 59.07  & 66.01 \\ 
 \Xhline{2\arrayrulewidth}

 \end{tabular}
 \label{table:ablation}
\end{table}

\textbf{Effects of layer weighting hyper-parameters}.
In this ablation study line, we investigate the effects of  $\alpha$ and $\beta_l$ on the final model performance. We tested with a wide range of $\alpha$ and $\beta_i$ using the grid-search strategy. Other hyper-parameters are kept the same as in our best-performing model. For the sake of simplicity, in Table \ref{alpha_beta_ablation}, we chose to show only the performance of our best-performing model ($2^{\text{nd}}$ row) with four representative models at four border settings of each hyper-parameter. From these results, we concluded that using $\alpha = 0.7$, $\beta_1 = 0.2 $, and $\beta_2 = 0.1$ yields the highest performance.
\begin{table}[t]
\centering
 \caption{Effect of the hyper-parameters $\alpha$ and $\beta$ on the tracking performance on the validation set. Here, the best and second-best methods for each case are highlighted using \textbf{bold} and \underline{underline}, respectively}
\setlength{\tabcolsep}{4mm}
 \begin{tabular}{c c c } 
 \Xhline{2\arrayrulewidth}
  \ Model name   & HOTA$\uparrow$ & IDF1$\uparrow$ \\ [0.5ex]
 \Xhline{2\arrayrulewidth}
   $\alpha = 1$, $\beta_1 = 0$, $\beta_2 = 0$  & \underline{52.57}  &  \underline{63.52} \\ 
  \textbf{$\alpha = 0.7$, $\beta_1 = 0.2$, $\beta_2 = 0.1$}  & \textbf{59.60}  & \textbf{66.39} \\ 
   $\alpha = 0.4$, $\beta_1 = 0.3$, $\beta_2 = 0.3$   & 44.95 & 47.37 \\
   $\alpha = 0$, $\beta_1 = 1$, $\beta_2 = 0$   & 51.34 & 54.57 \\
  $\alpha = 0$, $\beta_1 = 0$, $\beta_2 = 1$   & 43.08  & 44.55 \\
  
 \Xhline{2\arrayrulewidth}

 \end{tabular}
 \label{alpha_beta_ablation}
\end{table}
\subsubsection{Complexity Analysis (Q4)}
We empirically validate the average inference time of a single frame from a video sequence in each MOT dataset. It is worth noting that we could not reproduce every method due to the lack of publicly available source code. We record the running time of the inference phase in seconds and include the final results in the last column of Table \ref{table:experimental_results}. Unsurprisingly, methods based on traditional image processing such as SORT and V-IOU perform the best in terms of speed, as they do not use a deep learning model. Among deep learning-based methods, our method achieves the fastest, followed by UNS20regress, while the supervised model GCNNMatch performs longest among all competitive schemes. The large difference between inference times may due to differences in model sizes and optimizations. 

\section{Conclusions}
In this paper, we presented UnsMOT, a unified framework for unsupervised multi-object tracking. Compared with previous methods, our work is the first to utilize an unsupervised GNN together with a self-supervised CNN and RNN to incorporate intra-frame topology, appearance, and motion information for tracking. Especially, our novel GNN branch was optimized using unsupervised loss functions to extract topology guidance for refining detection-tracklet association, and it effectively boosts the tracking performance. UnsMOT showed state-of-the-art performance for unsupervised multi-object tracking problems on the prominent MOT challenge benchmarks. The performance of UnsMOT even outperformed those of many popular supervised techniques, confirming the ability to solve the tracking problem without costly data annotation efforts.

\appendices


\ifCLASSOPTIONcaptionsoff
  \newpage
\fi



%
\bibliographystyle{ieeetr}
\bibliography{egbib}
%

\begin{IEEEbiography}[{\includegraphics[width=1in,height=1.25in,clip,keepaspectratio]{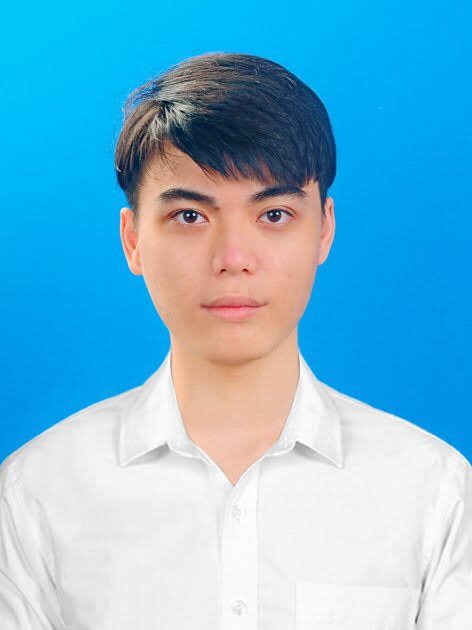}}]{Son Tran}  received his bachelor's degree in Information Technology from Post and Telecommunication Institute of Technology, Hanoi, Socialist Republic of Vietnam, in 2021. Since 2020, he has been an undergraduate researcher at the Faculty of Information Technology, Posts \& Telecommunication Institute of Technology, Hanoi, Vietnam. His research interests include graphical machine learning, deep learning, and computer vision.
\end{IEEEbiography}

\begin{IEEEbiography}[{\includegraphics[width=1in,height=1.25in,clip,keepaspectratio]{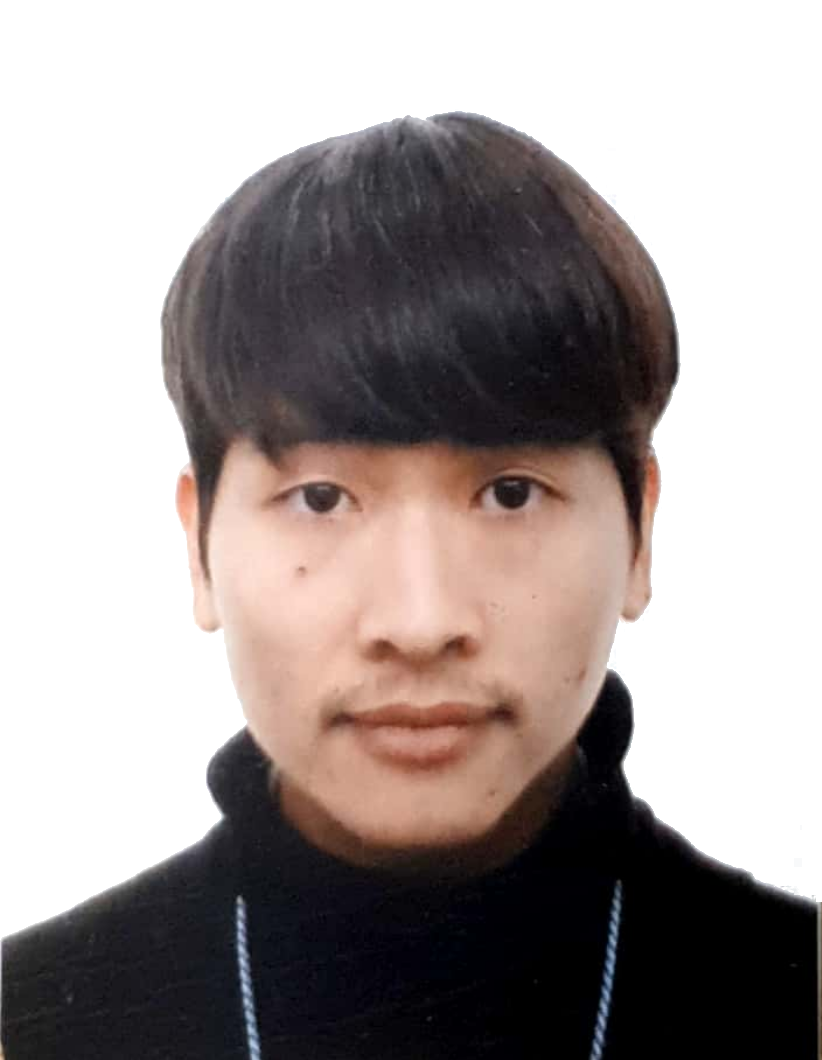}}]{Cong Tran}  received his doctoral degree in computer science from Dankook University, Yongin,
Republic of Korea, in 2021. He previously received his M.Sc. in computer science in 2014 and his B.Sc. in network and communication in 2009 from Vietnam National University, Hanoi, Vietnam. Since September 2021, he has been with the Faculty of Information Technology, Posts \& Telecommunication Institute of Technology, Hanoi, Vietnam, as a lecturer. His research interests include social network analysis, data mining, and machine learning.
\end{IEEEbiography}

\begin{IEEEbiography}[{\includegraphics[width=1in,height=1.25in,clip,keepaspectratio]{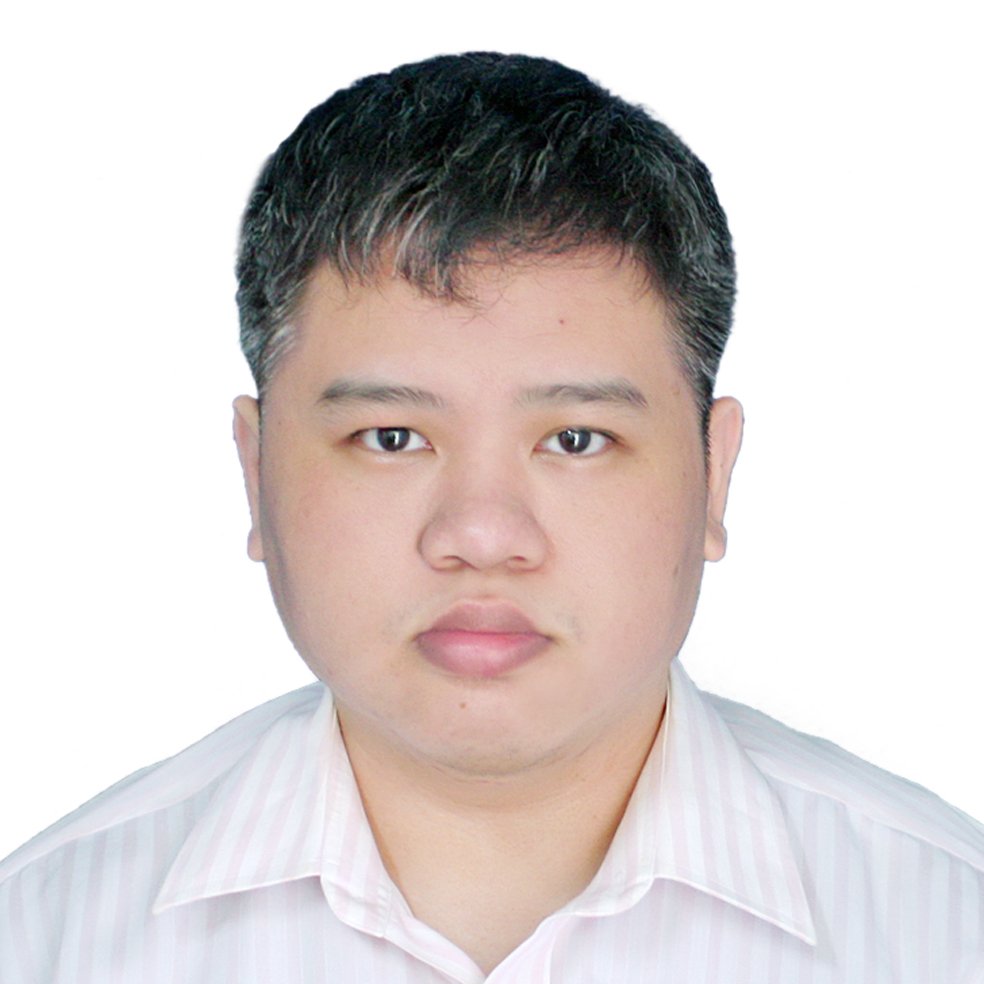}}]{Anh Tran} received his doctoral degree in Computer Science from University of Southern California in 2017, working with Professor Gerard Medioni. He previously received his B.Eng. degree from the Hanoi University of Science and Technology in 2010. In 2018-2019, he was an Applied Scientist at Amazon Rekognition, working on facial image processing APIs. Since August 2019, he has been a research scientist at VinAI Research. His research interests are in computer vision, particularly in human image analysis. He has received several honors including Vietnam Talents 2010, Imagine Cup Vietnam 2009, and Vietnam Education Foundation fellowship 2012.
\end{IEEEbiography}

\begin{IEEEbiography}[{\includegraphics[width=1in,height=1.25in,clip,keepaspectratio]{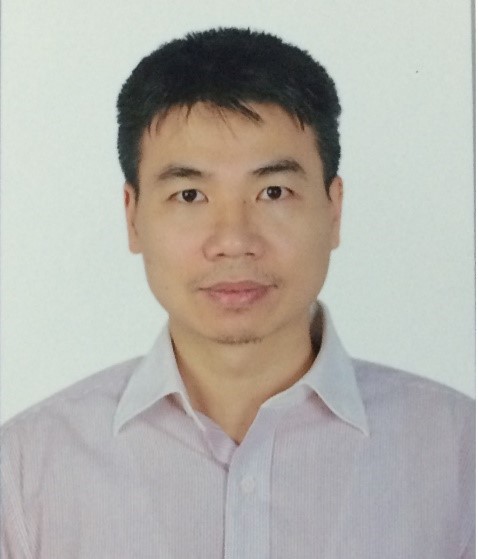}}] {Cuong Pham} received a PhD in Computer Science at Newcastle University in 2012. He is an Associate Professor of Computer Science and the Director of NAVER AI center at Posts and Telecommunications Institute of Technology. He is also a Visiting Research Scientist at VinAI Research. His research interests include machine learning/deep learning, ubiquitous computing, wearable computing, computer vision, human activity recognition, human computer interaction, and pervasive healthcare.
\end{IEEEbiography}




\end{document}